%% file: main.tex
\documentclass[conference]{IEEEtran}
\IEEEoverridecommandlockouts
\usepackage{cite}
\usepackage{amsmath,amssymb,amsfonts}
\usepackage{algorithmic}
\usepackage{graphicx}
\usepackage{textcomp}
\usepackage{xcolor}
\usepackage{multirow}
\usepackage{booktabs}
\usepackage{hyperref}
\usepackage[T1]{fontenc}

\def\BibTeX{{\rm B\kern-.05em{\sc i\kern-.025em b}\kern-.08em
    T\kern-.1667em\lower.7ex\hbox{E}\kern-.125emX}}
\begin{document}

\title{IPCGRL: Language-Instructed Reinforcement Learning for Procedural Level Generation}

\author{\IEEEauthorblockN{In-Chang Baek\textsuperscript{1}\IEEEauthorrefmark{2}, Sung-Hyun Kim\textsuperscript{1}\IEEEauthorrefmark{2}, Seo-Young Lee\textsuperscript{1}, Dong-Hyeon Kim\textsuperscript{2}, Kyung-Joong Kim\textsuperscript{1\IEEEauthorrefmark{3}}}
\IEEEauthorblockA{
\textit{\textsuperscript{1}Gwangju Institute of Science and Technology (GIST), South Korea} \\
\textit{\textsuperscript{2}Dongseo University, South Korea} \\
\{\texttt{inchang.baek, st4889ha}\}{@gm.gist.ac.kr} \\
}
\thanks{\IEEEauthorrefmark{2} Equal contribution \IEEEauthorrefmark{3} Corresponding author}
}

\maketitle

\begin{abstract}
\input{abstract/abstract.tex}
\end{abstract}


\section{Introduction}
\input{introduction/introduction}

\begin{figure}[!t]
    \centering
    \includegraphics[width=1.0\linewidth]{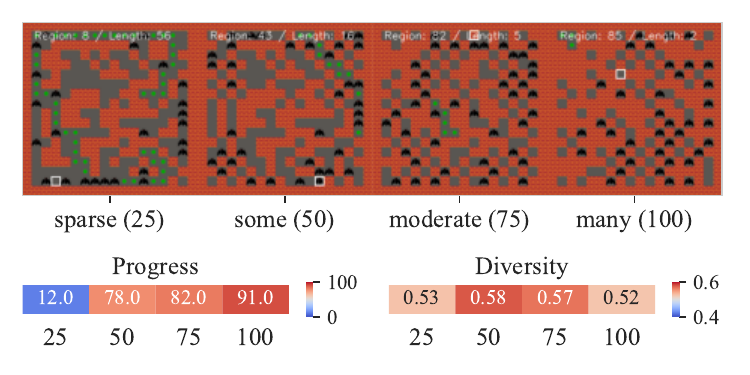}
    \vspace{-0.8cm}
    \caption{The single-task generation result with a single-task generator model trained on the \textit{RG} task ($\tau_{\scriptscriptstyle\text{RG}}$) instruction set and evaluated on the same instruction set. Rendered images: \textit{“Sparse regions”}, \textit{“Some regions”}, \textit{“Moderately scattered regions”}, and \textit{“Many regions”}.}
    \label{fig:single_region}
    \vspace{-0.5cm}
\end{figure}

\begin{figure}[!t]
    \centering
    \includegraphics[width=1.0\linewidth]{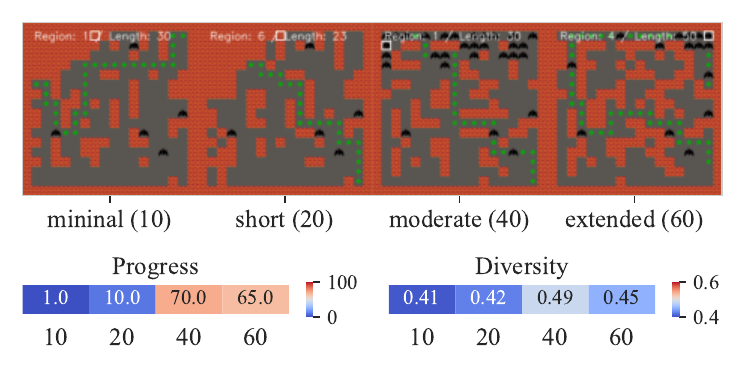}
    \vspace{-0.8cm}
    \caption{The single-task generation result with a single-task generator model trained on the \textit{PL} task ($\tau_{\scriptscriptstyle\text{PL}}$) instruction set and evaluated on the same instruction set. Rendered images: \textit{“Nano path length”}, \textit{“Short path length”}, \textit{“Moderate path length”}, and \textit{“Significant path length”}.}
    \label{fig:single_pl}
    \vspace{-0.5cm}
\end{figure}

\section{Related Works}

\input{background/background}

\begin{figure}[!t]
    \centering
    \includegraphics[width=1.0\linewidth]{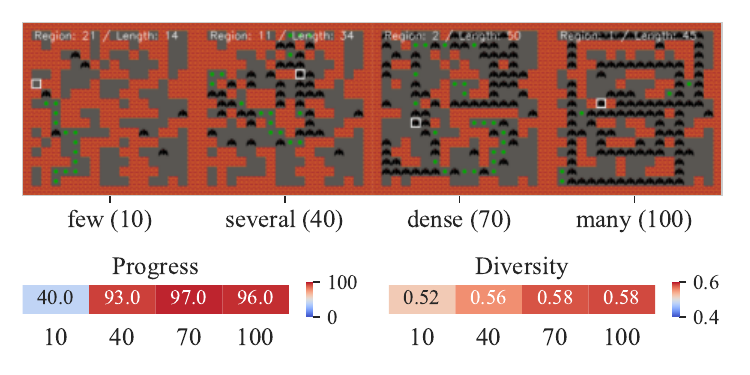}
    \vspace{-0.8cm}
    \caption{The single-task generation result with a single-task generator model trained on the \textit{BC} task ($\tau_{\scriptscriptstyle\text{BC}}$) instruction set and evaluated on the same instruction set. Rendered images: \textit{“Few bats”}, \textit{“Some bats”}, \textit{“Multi-bat cluster”}, and \textit{“Many bats”}.}
    \label{fig:single_bc}
    \vspace{-0.5cm}
\end{figure}

\section{Task Definition}
\input{experiment/environment}

\begin{figure}[!t]
    \centering
    \includegraphics[width=1.0\linewidth]{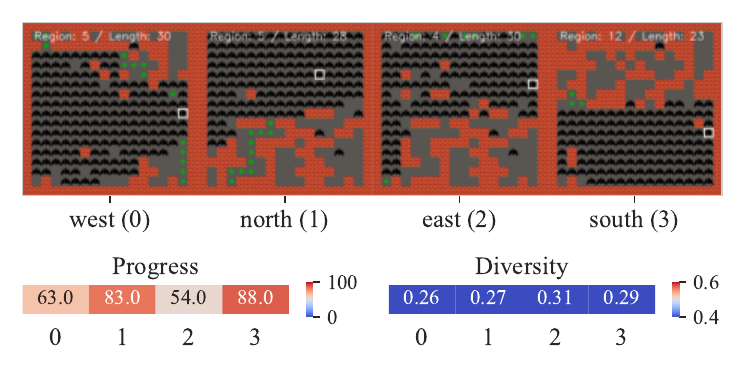}
    \vspace{-0.8cm}
    \caption{The single-task generation result with a single-task generator model trained on the \textit{BD} task ($\tau_{\scriptscriptstyle\text{BD}}$) instruction set and evaluated on the same instruction set. Rendered images: \textit{“Bats in the west”}, \textit{“Bats in the north”}, \textit{“Bats in the east”}, and \textit{“Bats in the south”}.}
    \label{fig:single_bd}
    \vspace{-0.5cm}
\end{figure}

\section{The IPCGRL Framework}

\input{method/method}

\section{Experiment Setup}
\input{experiment/setup}

\section{Experimental Result}

\input{experiment/experiment}

\section{Discussion}
\input{discussion/discussion}

\section{Conclusion and Future Work}
\input{conclusion/conclusion}

\section{Acknowledgments}
This research was supported by Culture, Sports and Tourism R\&D Program through the Korea Creative Content Agency grant funded by the Ministry of Culture, Sports and Tourism in 2025 (Project Number: RS-2024-00441523).

This research was also supported by Basic Science Research Program through the National Research Foundation of Korea (NRF) funded by the Ministry of Education (RS-2024-00395665).

\bibliographystyle{IEEEtran}
\bibliography{references}

\end{document}

%% file: abstract/abstract.tex
Recent research has highlighted the significance of natural language in enhancing the controllability of generative models. While various efforts have been made to leverage natural language for content generation, research on deep reinforcement learning (DRL) agents utilizing text-based instructions for procedural content generation remains limited.
In this paper, we propose \textit{IPCGRL}, an instruction-based procedural content generation method via reinforcement learning, which incorporates a sentence embedding model. IPCGRL fine-tunes task-specific embedding representations to effectively compress game-level conditions. We evaluate IPCGRL in a two-dimensional level generation task and compare its performance with a general-purpose embedding method.
The results indicate that IPCGRL achieves up to a 21.4\% improvement in controllability and a 17.2\% improvement in generalizability for unseen instructions with varied condition expressions within the same task. Furthermore, the proposed method extends the modality of conditional input, enabling a more flexible and expressive interaction framework for procedural content generation.

%% file: introduction/introduction.tex
Procedural content generation (PCG) has become an essential technique in game development and artificial intelligence, enabling the automatic creation of diverse, scalable, and adaptive game environments with minimal manual effort.
Several machine learning-based approaches have been developed using search-based methods, generative adversarial networks, and text generative models.
With recent advancements in deep reinforcement learning (DRL), it has become possible to train generative models in a data-free manner \cite{khalifa2020pcgrl}. The DRL approach optimizes an agent's policy using a reward function and introduces variations in the policy to achieve different objectives. This approach has shown promising results in content generation tasks.

Research on procedural content generation via reinforcement learning (PCGRL) has explored various topics, including numerically controllable content generation \cite{earle2021learning}, experience-driven generation \cite{mahmoudi2021arachnophobia}, applications in three-dimensional game environments \cite{jiang2022learning}, scalability issues \cite{earle2024scaling}, multiplayer content \cite{jeon2023raidenv} and reward function design methodologies \cite{baek2024chatpcg,baek2025pcgrllm}.
Typically, PCGRL methods define a specific goal as a reward function within reinforcement learning, allowing an RL agent to learn and generate content accordingly. However, since these methods rely on predefined numerical features, they struggle to adapt to complex or evolving objectives and have limited capacity to incorporate high-level human intent and intervention. Therefore, a more intuitive and flexible approach to content generation is needed that can overcome the constraints of traditional PCGRL.

\begin{figure}[!t]
    \centering
    \includegraphics[width=1.0\linewidth]{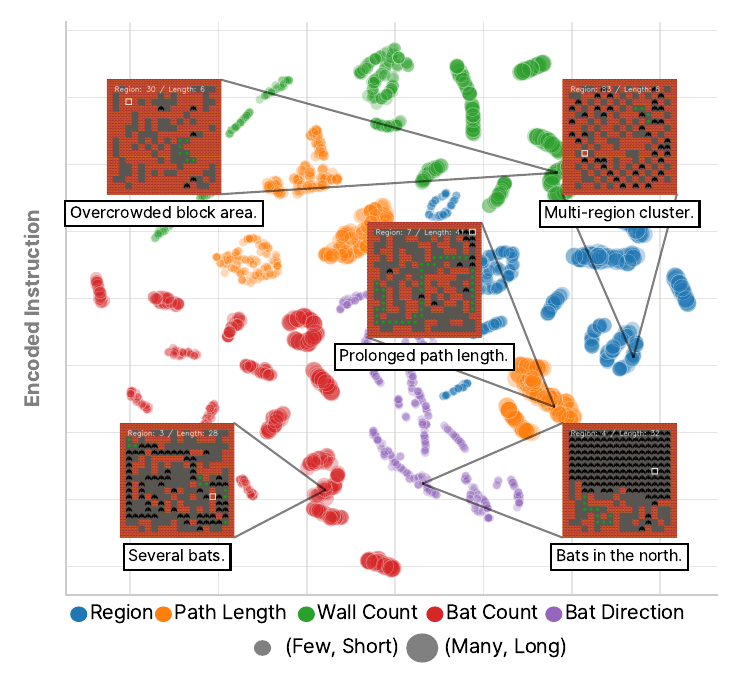}
    \vspace{-0.8cm}
    \caption{The encoded representation of instructions and the corresponding levels generated by IPCGRL agents. The color of the points represents the task, while their size, indicated by a gray circle, reflects the quantitative condition of the instructions.}
    \label{fig:concept}
    \vspace{-0.5cm}
\end{figure}

Natural language is a powerful tool for controlling DRL agents, as it conveys human knowledge intuitively and serves as an effective interface.
Language-instructed DRL enhances policy learning by leveraging natural language instructions, enabling agents to acquire goal-directed behaviors in domains such as robotics \cite{bing2023meta} and multi-agent cooperation \cite{hu2023language}.
However, despite these advantages, research on utilizing natural language as a conditioning input in RL-based content generation remains largely unexplored. 
Existing PCGRL approaches rely on predefined reward functions and manually designed constraints, which limit adaptability and expressiveness.
Leveraging language-instructed approaches, the model can be more flexibly controlled through a user-friendly modality.

To address these challenges, we introduce \textit{IPCGRL}, a language-instructed reinforcement learning framework tailored for text-to-level generation. IPCGRL integrates transformer-based sentence embeddings with an auxiliary encoder, guiding the model to effectively capture game-level features and conditions.
These encoded representations are provided during deep reinforcement learning (DRL) training, enabling the mapping of diverse condition expressions.
Furthermore, by incorporating natural language conditioning into PCGRL, IPCGRL facilitates more human-intuitive level generation while supporting flexible input formats.

To evaluate the effectiveness of IPCGRL, we conduct experiments on text-driven level generation tasks involving instructions of varying complexity. We assess the impact of embedding-based textual conditioning on generalization to unseen and compositional instruction sets. Benchmarking against a widely used two-dimensional level generation environment, IPCGRL demonstrates the capability to generate diverse level structures while maintaining precise control. Our results indicate that the proposed language-conditioned generative model enhances user accessibility without compromising performance compared to traditional controllable approaches. Furthermore, the use of text-based modality enables flexible control and expands user creativity.

This paper makes the following key contributions:
\begin{itemize} \item We propose a text-controlled PCGRL method that represents embedding vectors in a task-specific space.
\item We evaluate the task-specific embedding model against general-purpose embeddings and benchmark its controllability compared to traditional methods.
\item We assess the model’s generalizability in both single-task and multi-task settings and conduct an ablation study.

\end{itemize}

%% file: background/background.tex
\subsection{Controllable PCGRL}
PCGRL \cite{khalifa2020pcgrl}---a DRL-based content generation method---is a machine learning-based content generation method.
The benefits of PCGRL stem from its data-free nature and computational efficiency during inference, making it well suited for real-time content generation in games \cite{togelius2011search}.
In PCGRL, the level design process is framed as a Markov Decision Process (MDP), where level generation is learned through a trial-and-error approach.
At each step \( t \), the agent observes the game level as a state \( s_t \), selects an action \( a_t \) to modify a tile of the level, and transitions to a new state \( s_{t+1} \).
The agent then receives a reward: $r_t = R(s_t, s_{t+1})$, determined by a reward function ($R$) that evaluates the transition between states.

Controllable PCGRL is a technique to generate desired content with vectorized conditions. 
The authors of \cite{earle2021learning} first introduced the conditional DRL models in the two-dimensional level generation task.
The conditional observation (e.g., path length) is defined as $c=\text{sign}(g-f(s_{t}))$, where $g$ is the goal condition and $f(s_{t})$ is the current metric of each condition.
The conditional observation is concatenated to the observation of the one-hot level array.
The loss of metric is calculated to {$l_{t}=||g-f(s_{t})||_{L_{1}}$} and the agent gets a reward $r_{t}=l_{t-1}-l_{t}$.
The previous studies \cite{earle2021learning,jiang2022learning,jeon2023raidenv} employ numerical conditional observation that there are restrictions on input modality.

% Actual citation to be written tomorrow
\subsection{Language-Instructed DRL}
Language-instructed DRL is a methodology in which an agent learns to execute goal-directed behavior based on given instructions.
Unlike traditional reinforcement learning, where the agent receives only the environmental state as input, this approach provides both the environmental state and a natural language instruction.
This approach is widely utilized in various domains such as robotics \cite{bing2023meta} and multi-agent cooperation \cite{hu2023language}. It enables effective exploration and policy generalization in real-world environments--where rewards are sparse or complex--through the guidance of natural language instructions. Furthermore, language-instructed DRL aims to enhance the efficiency and adaptability of interactions by allowing the agent to directly understand and execute human linguistic directives.

There are two primary methods to leverage a given natural language instruction. The first approach is to input information extracted from the natural language instruction directly into the agent’s network; i.e., the instruction is used as direct input to the policy so that the agent can interpret the instruction and select appropriate actions. To achieve this, some studies convert the instruction into an intermediate language tailored to the task \cite{akakzia2020grounding, pang2023natural} or provide embeddings obtained via representation learning \cite{lynch2020language, sodhani2021multi, myers2023goal}. The second approach involves using the natural language instruction to induce a reward function, either by training a reward model based on the instruction \cite{bahdanau2018learning} or by designing an additional reward or regularization term that encourages the agent to learn a strategy corresponding to the instruction \cite{goyal2019using, hu2023language}, thus guiding the policy to maximize that reward. In addition, there exists a model-based approach that learns a language-instructed world model and solves tasks through planning\cite{dainese2023reader}.

%% file: experiment/environment.tex
\subsection{Environment}

This study employs a popular two-dimensional level generation gym environment \cite{khalifa2020pcgrl,earle2021learning} and we select the GPU-accelerated version \cite{earle2024scaling}.
The modified \textit{Dungeon} problem features three game tiles: \textit{Empty} \includegraphics[height=0.8em]{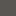}, \textit{Wall} \includegraphics[height=0.8em]{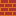}, and \textit{Bat} \includegraphics[height=0.8em]{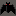};  we use the \textit{Turtle} representation, which the action space incorporates relative agent movement and tile replacement.
The observation comprises the positions of each tile type, along with the location of the tile to be modified.
The discrete action space includes seven behaviors: four actions correspond to moving the agent to the modification location, while the remaining three determine the specific tile type that replaces the tile at the modification location.
The level size was set to \( 16 \times 16 \), and the episode length is configured to 1,500 steps per episode.
Each episode begins with a randomly initialized matrix derived from the three-tile set.

\subsection{Conditional Generation Tasks}
\label{sec:cond_gen_task}
We define five tasks $\tau_i$ for usable controllable generation in the environment: region, path length, wall count, bat count, and bat direction.
\textit{\textbf{Region (RG)}} controls the number of independent regions in the level, where each region is a contiguous area separated from others by walls.
\textit{\textbf{Path Length (PL)}} controls the target distance between any two reachable points within the independent region.
\textit{\textbf{Wall Count (WC)}} controls the target number of wall tiles in the level.
\textit{\textbf{Bat Count (BC)}} controls the target number of bat tiles placed in the level.
\textit{\textbf{Bat Direction (BD)}} controls the distribution of bat tiles across the four cardinal directions. This task ensures that bat tiles are placed with specified directional orientations.
The first four tasks---\textit{RG}, \textit{PL}, \textit{WC}, and \textit{BC}---are conditioned on numerical target values 
$c$, while the \textit{BD} task is conditioned on one of four discrete directional values.

%% file: method/method.tex
\begin{figure*}[!t]
    \centering
    \includegraphics[width=0.90\linewidth]{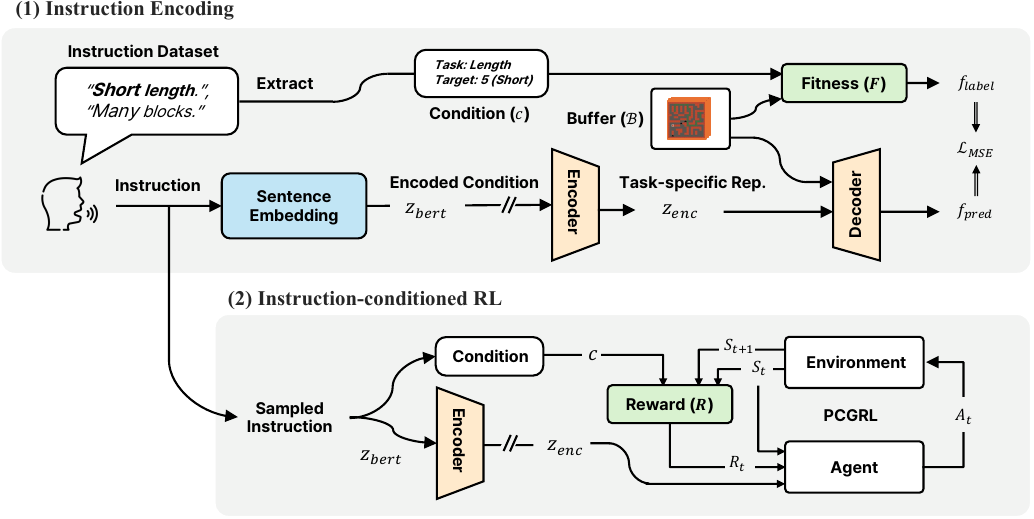}
    \caption{The IPCGRL framework consists of two main phases: (1) training the instruction encoder, and (2) training the DRL agent conditioned on the encoded instructions.
    The encoder is trained once with all task instructions and then used during the RL training process for each task.
    Details of each phase are provided in Section \ref{sec:encoder} and Section \ref{sec:conditional_pcgrl}, respectively. Dashed lines indicate gradient stop.}
    \label{fig:architecture}
    \vspace{-0.5cm}
\end{figure*}

We propose IPCGRL, a language-instructed PCGRL framework designed to train a language-instructed DRL agent for procedural level generation.
Fig. \ref{fig:architecture} illustrates the two-phase training process to develop a language-instructed PCGRL agent.
In the first phase, the encoder model is trained, and the trained encoder is then frozen during the training of the DRL agent in the second phase.
The following sections describe the procedure for collecting instruction data (Section \ref{sec:instruct_dataset}), training a language encoder model to compress conditional embeddings (Section \ref{sec:encoder}), and training a DRL agent using compressed embedding features (Section \ref{sec:conditional_pcgrl}).
The source code and the instruction sets are available in this repository\footnote{\url{https://github.com/bic4907/language-instructed-pcgrl}}.

\subsection{Instruction Dataset}
\label{sec:instruct_dataset}

\newcommand{\concat}{\oplus}

We construct a dataset of natural language instructions for both single-task and multi-task scenarios, corresponding to the five tasks defined in Section~\ref{sec:cond_gen_task}.
For the \textbf{single-task instructions} (e.g., $\tau_{\scriptscriptstyle\text{RG}}$), we generate a total of $80$ instructions, each describing a single task condition in natural language  (29.7$\pm{}$21.10 characters). 
The instruction set includes an empirically defined condition range. Each task is associated with a specific quantitative condition, such as the number of regions, path length, or the count of certain entities. These numerical targets are denoted as $c_{\scriptscriptstyle\text{RG}}, c_{\scriptscriptstyle\text{PL}}, c_{\scriptscriptstyle\text{WC}}, \ldots$, where each $c_i$ represents the target value  a specific predefined condition. To construct \textbf{multi-task instructions} (e.g., $\tau_{\scriptscriptstyle\text{RG}\concat{}\text{PL}}$), we combine pairs of single-task instructions, resulting in a total of $256$ multi-task instructions (61.1$\pm{}$32.08 characters). Each multi-task instruction describes two conditions simultaneously, representing more complex task requirements. The combined instructions are generated to ensure natural coherence between the two conditions, preserving logical consistency when merging the two descriptions. Each natural language instruction $\mathcal{I}$ can be interpreted as a set of one or more task–condition pairs $(\tau_i, c_i)$. 
To formalize this interpretation, we define a mapping function $\mathcal{C}$ that translates $\mathcal{I}$ into its corresponding set of task–condition pairs, such that $\mathcal{C}(\mathcal{I})=\{(\tau_{\scriptscriptstyle\text{RG}},c_{\scriptscriptstyle\text{RG}}),...,(\tau_{\scriptscriptstyle\text{BD}},c_{\scriptscriptstyle\text{BD}})\}$.

Table~\ref{tab:instruction} presents examples from the instruction dataset, covering both single-task and multi-task instructions, along with the corresponding numerical targets for each condition.
The mappings between qualitative expressions (e.g., \textit{few}, \textit{many}, \textit{significant}) and their corresponding numerical values. This approach ensures that the qualitative terms are consistent and contextually grounded in the task domain. For example, the expression \textit{multi-region cluster} corresponds to 75 regions, while \textit{overcrowded block area} corresponds to 160 blocks.

\begin{table}[!h]
    \centering

    \caption{The example of language instruction set}
    \begin{tabular}{p{0.6cm}p{4.5cm}p{0.15cm}p{0.15cm}p{0.15cm}p{0.15cm}p{0.15cm}}
    \toprule
    Task & Instruction & $c_{\scriptscriptstyle\text{RG}}$ & $c_{\scriptscriptstyle\text{PL}}$ & $c_{\scriptscriptstyle\text{WC}}$ & $c_{\scriptscriptstyle\text{BC}}$ & $c_{\scriptscriptstyle\text{BD}}$ \\
    \midrule
    $\tau_{\scriptscriptstyle\text{RG}}$ & \textit{A few}-regions are present. & 25 &  &  &  &  \\
    $\tau_{\scriptscriptstyle\text{RG}}$ & \textit{Multi}-region cluster. & 75 &  &  &  &  \\
    $\tau_{\scriptscriptstyle \text{PL}}$ & \text{Nano} path length. &  & 10 &  &  &  \\
    $\tau_{\scriptscriptstyle \text{PL}}$ & The path is \textit{extended} and \textit{prolonged}, requiring considerable traversal effort. &  & 60 &  &  &  \\
    $\tau_{\scriptscriptstyle \text{WC}}$ & \textit{A few} blocks are scattered \textit{sparingly}. &  &  & 10 & &  \\
    $\tau_{\scriptscriptstyle \text{WC}}$ & 
    \text{Overcrowded} block area. &  &  & 160 &  &  \\
    $\tau_{\scriptscriptstyle \text{BC}}$ & \textit{A Few} bats are scattered across the map. &  &  &  & 10 &  \\
    $\tau_{\scriptscriptstyle \text{BD}}$ & Bats spread at the \textit{bottom}. &  &  &  &  & 3 \\
    $\tau_{\scriptscriptstyle \text{BD}}$ & Bats in the \textit{north}. &  &  &  &  & 1 \\
    $\tau_{\scriptscriptstyle \text{PL}\concat{}\text{BC}}$ & \textit{Extended} path length, a \textit{substantial number} of monsters are scattered across the map. &  & 60 &  & 50 &  \\
    % $\tau_{\scriptscriptstyle \text{PL}\concat{}\text{BD}}$ & Significant path length, The bats are gathered in the north, heavily occupying the top section. &  & 60 &  &  & 1 \\
    % $\tau_{\scriptscriptstyle \text{WC}\concat{}\text{BD}}$ & Overcrowded block area, Top-focused bat distribution. &  &  & 160 &  & 1 \\
    % $\tau_{\scriptscriptstyle \text{BC}\concat{}\text{BD}}$ & Few bats, Bats grouped at the top. &  &  &  & 10 & 1 \\
    \bottomrule
    \end{tabular}
    \label{tab:instruction}
\end{table}

\subsection{Conditional Instruction Encoder}
\label{sec:encoder}
The objective of the training instruction encoder proposed in this study is to generate an embedding vector for a given natural language instruction $\mathcal{I}$. 
This embedding, denoted as $z_{enc}$, is a 64-dimensional vector that captures the semantic representation of the instruction.
The encoder generates the embedding vector $z_{enc}=E_{\theta}(z_{bert})$, where the input vector is obtained as $z_{bert} = BERT_{\phi}(\mathcal{I})$.
Here, $\phi$ denotes the fixed pretrained parameters of the BERT model.
The decoder is trained to predict the fitness value as $f_{pred}=D_{\theta}(s',z_{enc}), s' \sim \mathcal{B}$, where state $s'$ is sampled from the offline state dataset $\mathcal{B}$.
The label fitness value is computed using the fitness function as follows Eq. \ref{eq:fitness}.

\begin{equation}
\label{eq:fitness}
f_{label}=F_{\tau}(s',c_i),  s'\sim \mathcal{B}, c_i \in \mathcal{C}(\mathcal{I})
\end{equation}

The fitness function evaluates how well a given state $s'$ satisfies a single condition $c_i$ contained in the natural language instruction $\mathcal{I}$.
For example, given the instruction "\textit{Overcrowded block area (160)}", the fitness function computes the difference between the number of walls in the state $s'$ and the condition value of $c_{\scriptscriptstyle\text{WC}}=160$.

Given a natural language instruction $\mathcal{I}$, the network first converts the instruction into a vector representation $z_{bert}$ using a pre-trained bidirectional encoder representation from transformers (BERT) \cite{devlin2019bert} model.
The task-specific representation is then generated by passing $z_{bert}$ through the encoder $E_\theta$. 
The pre-trained BERT model remains frozen during training, while the encoder $E_\theta$ is implemented with a model with two dense layers.
The encoder model serves two primary functions:

\begin{itemize}
\item Dimensionality reduction: Compresses the high-dimensional representation into a low-dimensional one, improving the efficiency and stability of representation learning.
\item Task-specific representation: Transforms a general-purpose representation into a domain-specific one suitable for level generation tasks.
\end{itemize}

The input state $s'$ is processed through a single convolutional layer before being passed into the decoder $D_\theta$ .
The offline state dataset $\mathcal{B}$ consists of all trajectories collected during the training process of the controllable PCGRL \cite{earle2024scaling} baseline model for each task. 
This dataset enables the decoder to learn over a diverse set of $(s', z_{enc})$ pairs, ensuring that the fitness function's task-specific representation is properly embedded into the instruction encoding.

The encoder-decoder model is trained to minimize the difference between the predicted fitness value $f_{pred}$ and the ground-truth fitness value $f_{label}$. 
This is achieved by optimizing the following mean squared error (MSE) loss function described on Eq. \ref{eq:mse}.

\begin{equation}
\label{eq:mse}
\mathcal{L}=\mathbb{E}_{s' \sim \mathcal{B}, c_i \in \mathcal{C}(\mathcal{I})}[(f_{label}-f_{pred})^2]
\end{equation}
The ultimate goal of the training process is not to optimize the decoder’s performance, but to enable the encoder to learn a meaningful task-specific representation. 
The decoder is solely used as an auxiliary training component and is discarded after training is completed.
The trained encoder is designed to extract only the essential information required to reconstruct the fitness function from the input instruction. 
Accordingly, it effectively compresses and filters the instruction into a representation that captures task-relevant information.
This ensures that only the most critical aspects of the natural language instruction are passed to the DRL policy, enabling direct utilization of natural language instructions in reinforcement learning environments.

% 간단하게 뉴럴넷 아키텍처 언급

\subsection{Instruction-conditioned PCGRL}
\label{sec:conditional_pcgrl}

The language-instructed DRL agent policy, denoted as $\pi{}(a|s,\mathcal{I})$, is trained using the sentence embedding $z$.
In each episode, a set of instructions ($\mathcal{I}$), equal in number to the environments, is uniformly sampled from the instruction dataset within the selected task.
The DRL policy samples an action $a_{t}=\pi(s_{t},z_{enc})$ based on the concatenated input $\{s_{t}, z_{enc}\}$, where $s_{t}$ denotes the environmental state at time step $t$.
The next environmental state $s_{t+1}$ is deterministically derived using the transition function 
$s_{t+1}=f(s_{t},a_{t})$.
The reward is calculated with the transition and the instruction condition $r_{t}=R(s_{t}, s_{t+1}, \mathcal{C}(\mathcal{I}))$.
The multi-conditioned reward function is formally defined as Eq. \ref{eq:reward}.

\begin{equation}
\label{eq:reward}
R(s_{t}, s_{t+1}, \mathcal{C}(\mathcal{I})) = \sum_{i=1}^{n} R_{\tau_i}(s_{t}, s_{t+1}, c_{i}) \cdot w_{\tau_i}
\end{equation}

\begin{figure*}[!t]
    \centering
    \includegraphics[width=1.0\linewidth]{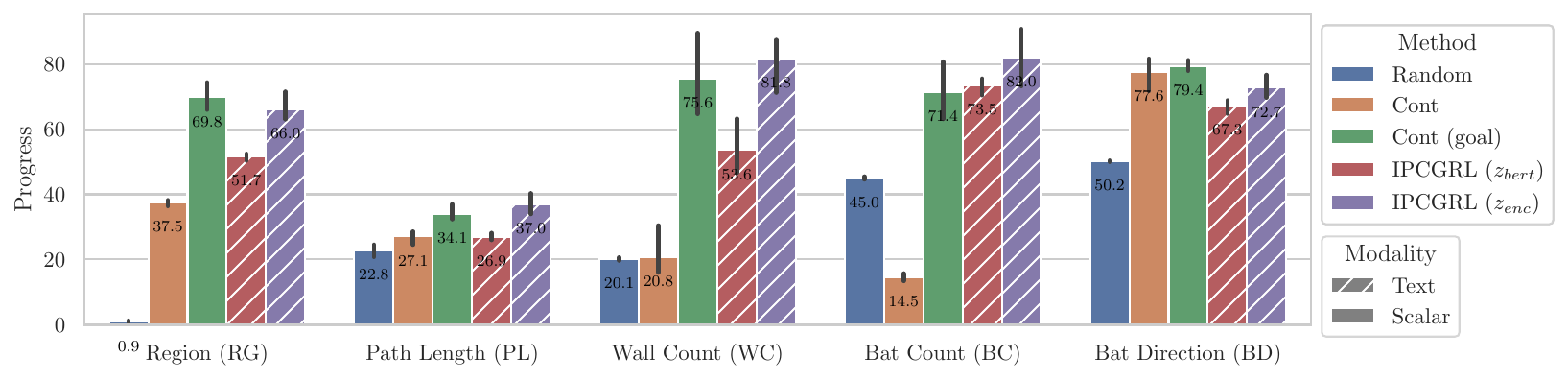}
    \vspace{-0.8cm}
    \caption{The comprehensive results of single-task generation across five task domains show that our proposed method, IPCGRL ($z_{enc}$), outperforms the model without task-specific representations ($z_{enc}$) in the text-control domain. IPCGRL achieves performance comparable to scalar-controlled generators (Cont) when provided with textual inputs.}
    \label{fig:single_task}
    \vspace{-0.5cm}
\end{figure*}

The reward is computed as the weighted sum of individual sub-task reward functions $R_{\tau_i}$, where each sub-task reward corresponds to condition $c_i$ and is weighted by $w_{\tau_i}$.
The sub-task functions such as $R_{\tau_{\scriptscriptstyle\text{RG}}}$ and $R_{\tau_{\scriptscriptstyle\text{PL}}}$ are loss-based controllable reward functions designed to encourage the generation to meet the specific conditions.
The weight $w_{\tau_{i}}$ for each task is used to scale the magnitude of its corresponding reward function. 
The parameter $n$ denotes the number of tasks included in the language instruction $\mathcal{I}$.
For single-task instructions, $n=1$, while for multi-task instructions, $n \geq 2$.

The policy network architecture processes the environmental state and the language instruction in a two-stream manner.
The environmental state $s_{t}$, represented as a one-hot matrix, is first encoded into a feature vector using two convolutional layers, followed by a flattening operation.
The final input to both the policy and value networks is formed by concatenating the flattened environmental state feature with the encoded instruction embedding.
This concatenated feature vector is used as input to both the policy and value networks, allowing the agent to condition its decisions on both the current environmental state and the provided language instruction.

%% file: experiment/setup.tex
\subsection{Model Training Setting}
We employ the pre-trained BERT model \cite{devlin2019bert} as the sentence embedding model, and stack two dense layers for both the encoder and decoder networks. The BERT output vector $z_{\text{bert}} \in \mathbb{R}^{768}$ is projected into a latent space $z_{\text{enc}} \in \mathbb{R}^{64}$ through the encoder network.
We collect a state buffer of 86K samples using the controllable PCGRL agent \cite{earle2024scaling} to cover a diverse distribution of states, excluding duplicate state data from the dataset.
Since the RL policy takes both the state and the condition as input, it is crucial to learn an embedding representation that generalizes well to the state distribution.
The encoder and decoder models are trained for 100 epochs, and the fitness of the tasks is normalized to the range [-5, 5].

The DRL models are trained using proximal policy optimization (PPO) \cite{schulman2017proximal} for a total of 50 million timesteps, using the \textit{PureJaxRL} implementation \cite{lu2022discovered}.
The chosen timesteps setting ensures sufficient convergence time in both single- and multi-task experiments.
The hyperparameters are set as follows: $10$ epoch size, $128$ rollout length, $\gamma_{\text{GAE}} = 0.95$, $\gamma = 0.99$, $10^{-4}$ learning rate.
Each comparison is evaluated over three independent runs, and the results are averaged. All experiments are conducted on machines equipped with a RTX 8000 GPU.
The weights of the reward functions are set to \( w_{PL} = 1 \) and \( w_{BC} = 0.083 \), considering reward magnitude and sparsity for the multi-task experiment.

\subsection{Evaluation Criteria}
The evaluation metrics for \textit{Progress} and \textit{Diversity} follow the definitions proposed in \cite{earle2021learning}. Each metric is derived by averaging the values across 10 levels generated per instruction, obtained through inference with the trained DRL models.

\textbf{Progress ($P$)} measures the relative percentage change from the initial state toward the specified target conditions. We improved the progress function to account for cases where the agent's generated outcome exceeded the goal, leading to increased loss: $1 - \lvert{ \frac{g - s_{T} }{g -s_0 }}\lvert$ where $s_T$ is the terminal state, and the loss is normalized with respect to the initial state $s_0$.
This function measures the progress of modifications toward achieving the goal state using a normalized score, which is scaled by 100 to produce a value within the range $[0, 100]$. Additionally, it considers scenarios where exceeding the target results in inefficiencies or unintended negative consequences.

\textbf{Diversity} measures as the mean per-level Hamming distance, computed by first calculating the average per-tile Hamming distance for each generated level and then taking the mean across all levels for an instruction. This value is normalized by the total map size, ensuring that the diversity score falls within the range \([0,1]\), where 0 indicates all maps are identical, and 1 indicates that all maps are entirely different.

\subsection{Comparison}

To the best of our knowledge, there is no language-instructed PCGRL agent, so we compare our approach with vector-based controllable agent models. \textbf{Cont}, conventional controllable PCGRL \cite{earle2021learning}, uses a condition of the form \( c = \text{sign}(g - f(s_t)) \) to provide directional control. However, the directional information is limited to the range \([-1,1]\), resulting in information loss about the target condition and state imbalance from the initial state.
To ensure a fair comparison, we developed a modified version of Cont, termed \textbf{Cont (goal)}, where the goal \( c_t=g \) is used directly as the condition for equal conditional input with our method. These models rely on numerical condition inputs and serve as baselines to benchmark the performance of language-conditioned DRL.
Our model, \textbf{IPCGRL (\( z \))}, is a natural language-based conditional generation approach. It has two variants: \textbf{\( z_{bert} \)}, which uses BERT-based embeddings, and \textbf{\( z_{enc} \)}, which incorporates task-specific representation features.
\textbf{Random} agent uniformly selects an action regardless of the state.

%% file: experiment/experiment.tex
\subsection{Single-Task Generation with Encoded Latent Space}
\label{exp:singletask}
\input{experiment/singletask}

\subsection{Generalizability on Unseen Instruction Set}
\input{experiment/generalizability}

\subsection{Generalizability on Multi-Task Generation}
\label{sec:multitask}
\input{experiment/multitask}

%% file: experiment/singletask.tex
We assess the controllability of different methods by using the scalar-controlled generator as a reference point and primarily focus on comparing text-modality approaches. Fig. \ref{fig:single_task} illustrates the performance progress of five single-task instructions (Section \ref{sec:instruct_dataset}), highlighting the differences between text- and scalar-conditioned generators. The generated levels are illustrated in Fig. \ref{fig:single_region}–\ref{fig:single_bd}.
To compare the embedding representation capability within the same modality, we compute an improvement metric by normalizing the performance of the proposed IPCGRL ($z_{\text{enc}}$) with respect to the baseline IPCGRL ($z_{\text{bert}}$).

For instruction-controlled generators, the IPCGRL ($z_{enc}$) model demonstrates an average performance improvement of 21.4\% over the IPCGRL ($z_{bert}$) model.
Specifically, IPCGRL ($z_{enc}$) achieves significant performance improvements of 21.6\%, 27.2\%, and 34.4\% in the RG, PL, and BC tasks compared to the IPCGRL ($z_{bert}$) model, respectively ($p<0.05$), as well as 11.44\% and 12.37\% improvements in the BC and BD tasks, respectively.
These results suggest that task-specific embeddings effectively compress the latent space by attending to both conditional and numeric words, leading to significantly higher performance gains compared to a general-purpose sentence embedding space.

Furthermore, no significant performance degradation ($p \geq 0.05$) is observed for scalar-controlled generators (Cont (goal)), except in the BD task ($p \approx 0.04$). These findings indicate that IPCGRL exhibits comparable controllability to traditional controllable PCGRL agents across different input modalities. This suggests that IPCGRL can serve as a viable alternative for traditional scalar-controlled PCGRL methods, offering flexibility across various control mechanisms.

%% file: experiment/generalizability.tex
\vspace{-0.2cm}
\begin{table}[h]
\caption{Comprehensive results for generalizability}
\label{tab:generalizability}
\resizebox{\linewidth}{!}{
\begin{tabular}{lr|r|rrrr}
\toprule
 & In-Dist & Near-OOD & \multicolumn{4}{c}{Far-OOD} \\
 & $\tau_{\text{PL}}$ & $\tau_{\text{PL}}$ & $\tau_{\text{RG}}$ & $\tau_{\text{WC}}$ & $\tau_{\text{BC}}$ & $\tau_{\text{BD}}$ \\
\midrule
Random & 21.94 & 25.43 & \textbf{0.87} & 20.09 & \textbf{45.04} & 50.20 \\
Cont & 18.52 & -\phantom{10} & 0.00 & 30.92 & 30.08 & \textbf{50.21} \\
Cont (goal) & 22.79 & -\phantom{10} & 0.00 & 32.06 & 32.54 & 50.03 \\
IPCGRL ($z_{bert}$) & 22.71 & 39.65 & 0.00 & \textbf{35.71} & 25.93 & 50.03 \\
IPCGRL ($z_{enc}$) & \textbf{33.88} & \textbf{46.47} & 0.00 & 29.36 & 15.88 & 50.07 \\
\bottomrule
\end{tabular}
}
\end{table}

The robustness to unseen instruction sentences is crucial for ensuring the generalizability of text-conditioned generative models. We evaluate model performance across three different instruction sets in the instruction dataset: training instructions within the same task (in-distribution, in-dist), test instructions within the same task but unseen during training (near out-of-distribution, near-OOD), and instructions from entirely different tasks (far-OOD). We construct a near-OOD test set comprising 25\% of the number of training instructions by modifying condition values through the replacement of quantifier words (e.g., many). The average cosine similarity measured with BERT embeddings is $0.831 \pm 0.084$ between the train- and test instructions, indicating that the test set retains task semantics while introducing meaningful lexical variation.
We conduct the experiment on the PL task since its path structure is similar to the wall task (WC), but it uses different tile names, such as bat (BC, BD), and it provides the most neutral instruction among them.

Table \ref{tab:generalizability} presents a comparative analysis of performance progression on seen and unseen instructions for $\tau_{\scriptscriptstyle\text{PL}}$. The scalar-controlled models were excluded from the near-OOD evaluation as the instruction split was determined by uniqueness.
Although the three evaluation sets are similar, slight differences in their conditions necessitate comparing methods only within the same set to ensure fairness.
The IPCGRL ($z_{enc}$) model demonstrates superior performance compared to both the scalar-based and IPCGRL ($z_{bert}$) models under in-dist and near-OOD conditions, even outperforming scalar-controlled models in certain cases.
The two IPCGRL methods underperform compared to other methods on far-OOD tasks, showing little generalization to other words included in different tasks.
The task-specific latent representation ($z_{enc}$) achieves a 49.1\% improvement on seen instructions and a 17.2\% improvement on unseen instructions relative to the general-purpose embedding ($z_{bert}$). These findings indicate that task-specific embeddings enhance the model’s ability to generalize to unseen instructions within the same task domain.
Moreover, this suggests that IPCGRL is capable of handling diverse numerical expressions during inference.

%% file: experiment/multitask.tex
\begin{figure}[!t]
    \centering
    \includegraphics[width=\linewidth]{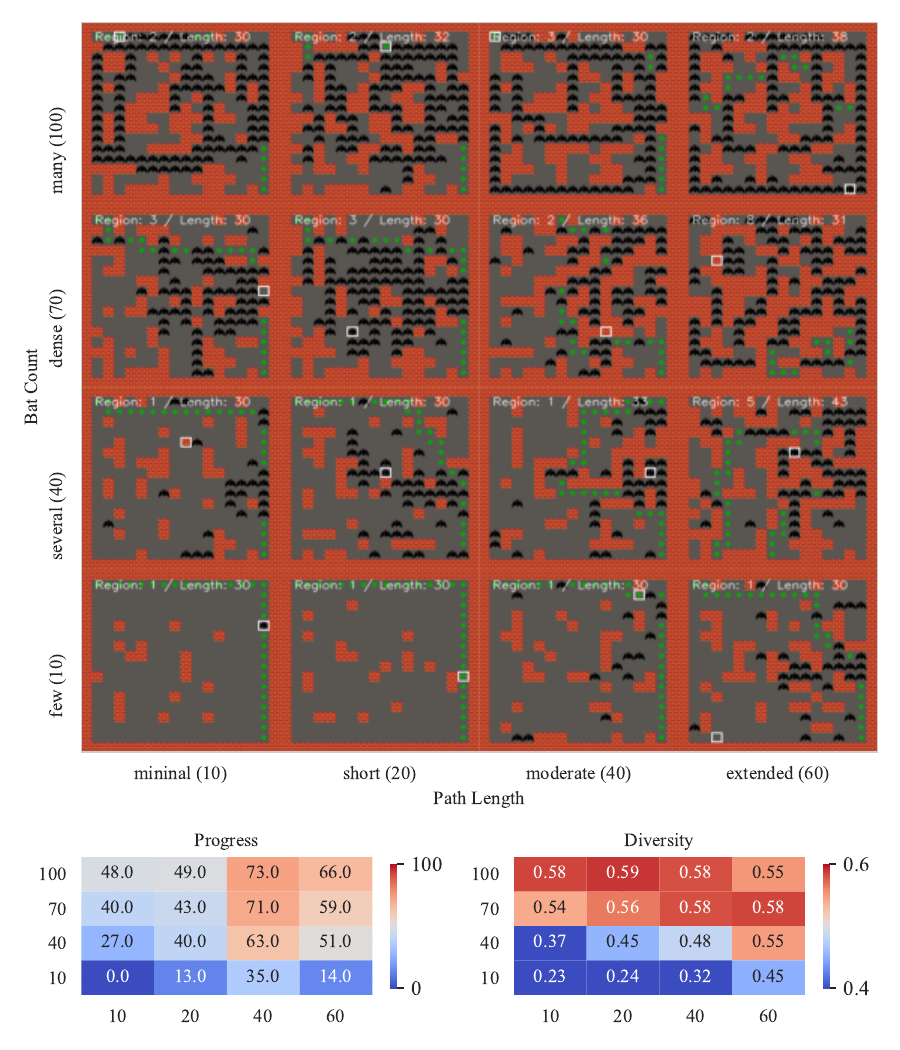}
    \vspace{-0.7cm}
    \caption{The multi-task generator trained on instructions consists of two tasks. The model is trained and evaluated with PL-BC combination instruction set ($\tau_{\scriptscriptstyle\text{PL}\oplus{}\text{BC}}$).  Under conditions with a large number of bats and a long map, it has been observed that the bats tend to generate longer diameter path. Instruction example: \textit{“\{Moderate\} path length, \{Many\} bats.”}}
    \label{fig:multitask}
    \vspace{-0.3cm}
\end{figure}

The multi-task generation experiment evaluates model performance across multiple tasks.  
Table \ref{tab:multitask} presents the instruction set $\tau_{\scriptscriptstyle\text{PL}+\text{BC}}$, which consists of two single-task instructions, $\tau_{\scriptscriptstyle\text{PL}}$ and $\tau_{\scriptscriptstyle\text{BC}}$, while $\tau_{\scriptscriptstyle\text{PL}\oplus\text{BC}}$ represents their textual combination. The visualization of multi-task generation is illustrated on Fig. \ref{fig:multitask}.
To assess generalizability, we benchmark the two IPCGRL models ($z_{bert}$, $z_{enc}$) on both single- and multi-task instruction sets, measuring their ability to generalize to both unseen multi-task and single-task instructions. 
For this experiment, we employ an encoder model trained on an instruction set consisting of single- and multi-task instructions.

\begin{table}[h]
\caption{Comprehensive result for multi-task generation}
\centering
\label{tab:multitask}
\begin{tabular}{p{0.7cm}p{2cm}p{0.8cm}p{0.8cm}p{0.8cm}p{0.8cm}}
\toprule
 & Test & \multicolumn{2}{c}{$\tau{}_{\scriptscriptstyle\text{PL}+\text{BC}}$} & \multicolumn{2}{c}{$\tau{}_{\scriptscriptstyle\text{PL}\oplus\text{BC}}$} \\
Train &  & $P_{\tau_{\scriptscriptstyle\text{PL}}}$ & $P_{\tau_{\scriptscriptstyle\text{BC}}}$ & $P_{\tau_{\scriptscriptstyle\text{PL}}}$ & $P_{\tau_{\scriptscriptstyle\text{BC}}}$ \\
\midrule
\multirow[t]{4}{*}{$\tau{}_{\scriptscriptstyle\text{PL}+\text{BC}}$} & IPCGRL ($z_{bert}$) & 22.61 & 59.67 & \textbf{21.48} & \textbf{47.41} \\
 & IPCGRL ($z_{enc}$) & \textbf{26.19} & \textbf{78.05} & 17.03 & 29.34 \\
\midrule
 \multirow[t]{4}{*}{$\tau{}_{\scriptscriptstyle\text{PL}\oplus\text{BC}}$}  & IPCGRL ($z_{bert}$) & \textbf{21.57} & 44.02 & \textbf{22.29} & \textbf{73.97} \\
 & IPCGRL ($z_{enc}$) & 20.96 & \textbf{49.71} & 21.64 & 65.93 \\
\midrule
\multirow[t]{4}{*}{$\tau{}_{\scriptscriptstyle\text{PL}+\text{BC}}$} & Cont & \textbf{28.66} & 12.23 & 18.95 & 13.40 \\
 & Cont (goal) & 26.27 & \textbf{77.32} & \textbf{23.02} & \textbf{59.61} \\
\midrule
\multirow[t]{4}{*}{$\tau{}_{\scriptscriptstyle\text{PL}\oplus\text{BC}}$} & Cont & 3.70 & 14.26 & 20.20 & 14.60 \\
 & Cont (goal) & \textbf{22.32} & \textbf{79.58}  & \textbf{28.07} & \textbf{80.05} \\
\bottomrule
\end{tabular}
\end{table}

To investigate the generalizability from single-task to multi-task learning, we trained the model using $r_{\scriptscriptstyle\text{PL}+\text{BC}}$, a row-concatenated instruction set comprising $r_{\scriptscriptstyle\text{PL}}$ and $r_{\scriptscriptstyle\text{BC}}$, and evaluated its performance on both $r_{\scriptscriptstyle\text{PL}+\text{BC}}$ and $\tau_{\scriptscriptstyle\text{PL}\oplus\text{BC}}$. 
The model trained on single-task instructions demonstrates superior performance on the corresponding single-task test set compared to the IPCGRL (\( z_{\text{bert}} \)) model; however, its performance degrades when evaluated on multi-task instruction sets.  
Furthermore, the IPCGRL (\( z_{\text{enc}} \)) underperforms relative to the IPCGRL (\( z_{\text{bert}} \)) model when trained on multi-task instructions and evaluated on single-task instructions.  
These findings suggest that while \( z_{\text{enc}} \) outperform in single-task settings with seen instructions, its capability to generalize during training and inference on multi-task instructions remains limited.

%% file: discussion/discussion.tex
\subsection{Ablation Study on Encoder Training}

\begin{figure}[!h]
    \centering

    \includegraphics[width=0.8\linewidth]{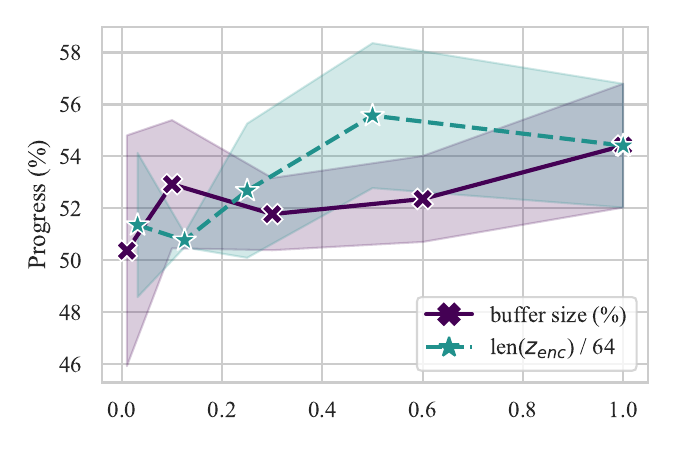}
    \vspace{-0.5cm}
    \caption{The performance correspond to the buffer size and embedding size.}
    \label{fig:ablation}
    \vspace{-0.5cm}
\end{figure}

In the ablation study (Fig. \ref{fig:ablation}), we examine the effects of buffer size ratio and encoder latent dimension (normalized by 64) on progress performance. The results indicate that smaller buffer sizes and shorter encoder lengths lead to a decline in performance. Notably, an appropriately sized encoder allows the model to effectively encode the complexity of the instruction set, enhancing representation learning while alleviating overfitting to train instruction set. Meanwhile, increasing the buffer size helps align the representation with diverse state distributions encountered during DRL training. These findings highlight that larger buffer sizes and appropriate encoder lengths contribute to improved robustness and generalizability.

\subsection{Multi-Task Generation Capability}
In multi-task generation, IPCGRL exhibits limited performance on training and testing with multi-task instruction set.
These results suggest that multi-task learning presents a significantly higher level of complexity compared to single-task learning.
One possible explanation is that the current encoder functions as a simple regression model with a single output node, which inherently limits its ability to explicitly distinguish between the tasks.
This finding highlights the need to consider a multi-label classification and regression structure to better capture and differentiate multiple task conditions.
This improvement is expected to enable the model to learn a smoothly interpolated embedding space, where a combination of unseen instructions is meaningfully positioned.

%% file: conclusion/conclusion.tex
In this study, we propose IPCGRL, a novel method that enables the PCGRL framework to incorporate text-level instructional conditions as input.
IPCGRL integrates a sentence embedding model and an encoder-decoder network to align the condition representation with the task-specific embedding space.
As a result, the proposed method achieves a 21.4\% performance improvement compared to PCGRL using pretrained BERT embeddings and demonstrates better generalization ability for unseen instructions within the same task domain.

Meanwhile, multi-task representation and reward normalization remain challenging issues that limit performance in multi-task generation.
To address this limitation, future work will incorporate a multi-label decoder to better capture multiple conditions.
Furthermore, extending the model to other modalities (e.g., vision) could enhance users' ability to express their intentions more naturally and intuitively.